\documentclass[manuscript,screen]{acmart}
\AtBeginDocument{%
  }

\definecolor{darkblue}{rgb}{0.0, 0.0, 0.55}

\begin{document}

\title{EAvatar: Expression-Aware Head Avatar Reconstruction with Generative Geometry Priors}


\author{Shikun Zhang}
\affiliation{%
  \institution{Department of Data Science and AI, Monash University}
  \city{Melbourne}
  \state{Victoria}
  \postcode{3800}
  \country{Australia}
}

\author{Cunjian Chen}

\affiliation{%
  \institution{Department of Data Science and AI, Monash University}
  \city{Melbourne}
  \state{Victoria}
  \postcode{3800}
  \country{Australia}
}

\author{Yiqun Wang}
\affiliation{%
  \institution{College of Computer Science, Chongqing University}
  \city{Chongqing}
  \postcode{401331}
  \country{China}
}


\author{Yong Li}
\affiliation{%
  \institution{College of Computer Science, Chongqing University}
  \city{Chongqing}
  \postcode{401331}
  \country{China}
}

\renewcommand{\shortauthors}{Zhang et al.}

\begin{abstract}
  High-fidelity head avatar reconstruction plays a crucial role in AR/VR, gaming, and multimedia content creation. Recent advances in 3D Gaussian Splatting (3DGS) have demonstrated effectiveness in modeling complex geometry with real-time rendering capability and are now widely used in high-fidelity head avatar reconstruction tasks. However, existing 3DGS-based methods still face significant challenges in capturing fine-grained facial expressions and preserving local texture continuity, especially in highly deformable regions. To mitigate these limitations, we propose a novel 3DGS-based framework termed EAvatar for head reconstruction that is both expression-aware and deformation-aware. Our method introduces a sparse expression control mechanism, where a small number of key Gaussians are used to influence the deformation of their neighboring Gaussians, enabling accurate modeling of local deformations and fine-scale texture transitions. Furthermore, we leverage high-quality 3D priors from pretrained generative models to provide a more reliable facial geometry, offering structural guidance that improves convergence stability and shape accuracy during training. Experimental results demonstrate that our method produces more accurate and visually coherent head reconstructions with improved expression controllability and detail fidelity. Project: https://saralkun.github.io/EAvatar.
\end{abstract}




\maketitle
\section{Introduction}
With the rapid advancement of VR/AR, visual effects, and game character generation technologies, high-quality and animatable 3D head avatar modeling has become a critical research topic in computer graphics and 3D vision fields~\cite{egger20203d,zollhofer2018state}.  In practical scenarios such as human-computer interaction~\cite{pantic2003toward}, real-time expression-driven animation~\cite{lombardi2018deep}, and personalized digital asset generation~\cite{grassal2022neural}, systems often require accurate modeling of head geometry and facial dynamics, along with fine-grained local control and real-time rendering capabilities. These demands pose significant challenges to existing 3D reconstruction methods, particularly in structural representation, detail fidelity, and deformation flexibility.
Current approaches to 3D face modeling can be broadly classified into three categories. 
The first category encompasses traditional mesh-based 3D Morphable Models (3DMMs), exemplified by the pioneering model developed by Blanz et al.~\cite{blanz2023morphable} and the more expressive FLAME model introduced by Li et al.~\cite{li2017learning}, which simultaneously models facial shape, expression, and pose. The second category comprises neural implicit function-based approaches, such as NeRF, NeuS~\cite{wang2021neus} and its derivatives tailored for dynamic face reconstruction (e.g., HyperNeRF~\cite{park2021hypernerf} and AvatarMe~\cite{grassal2022neural}). While these methods offer certain levels of personalization and expression control, they still suffer from several limitations. For instance, traditional 3DMMs can efficiently reconstruct facial geometry but often rely on global linear blend weights for expression control~\cite{booth2018large}, making it difficult to achieve fine-grained local edits~\cite{li2017learning}. Meanwhile, NeRF and NeuS-based methods are capable of modeling continuous implicit fields, but they often struggle with capturing high-frequency geometric details and maintaining local coherence in regions with significant deformations. These limitations become more pronounced in high-resolution settings or under extreme facial expressions, leading to texture blurring and geometric drifting artifacts~\cite{gafni2021nerface}.

Recently, 3D Gaussian Splatting (3DGS)~\cite{kerbl20233d} has emerged as a robust explicit representation for real-time rendering, modeling scenes through a collection of discrete 3D Gaussians while facilitating high-quality rasterization. 
As the third type of approach in head avatar reconstruction, this Gaussian-based strategy has been adopted by several recent works to achieve promising outcomes in head avatar reconstruction. For instance, HeadGaS~\cite{dhamo2024headgas} and PointAvatar~\cite{zheng2023pointavatar} both exploit dynamic Gaussians to build animatable head avatars with high rendering efficiency and realism. HeadGaS models expression changes using a global linear blending of parameters, which provides fast animation performance. However, it lacks fine-grained control over local areas. In contrast, PointAvatar relies on preset facial models, resulting in poor generalization when handling extreme expressions. To address coarse expression modeling, GaussianAvatars~\cite{qian2024gaussianavatars} combines Gaussians with a parametric face model to improve global expression controllability. Nevertheless, its reliance on mesh-based priors limits the representation of out-of-distribution local geometry. Another recent approach, SplattingAvatar~\cite{shao2024splattingavatar}, improves local detail rendering by embedding Gaussians into a deformable mesh structure, but its expressiveness is still constrained by the mesh’s low-frequency motion, especially under extreme expressions.

\begin{figure*}
\centering
  \includegraphics[width=\textwidth]{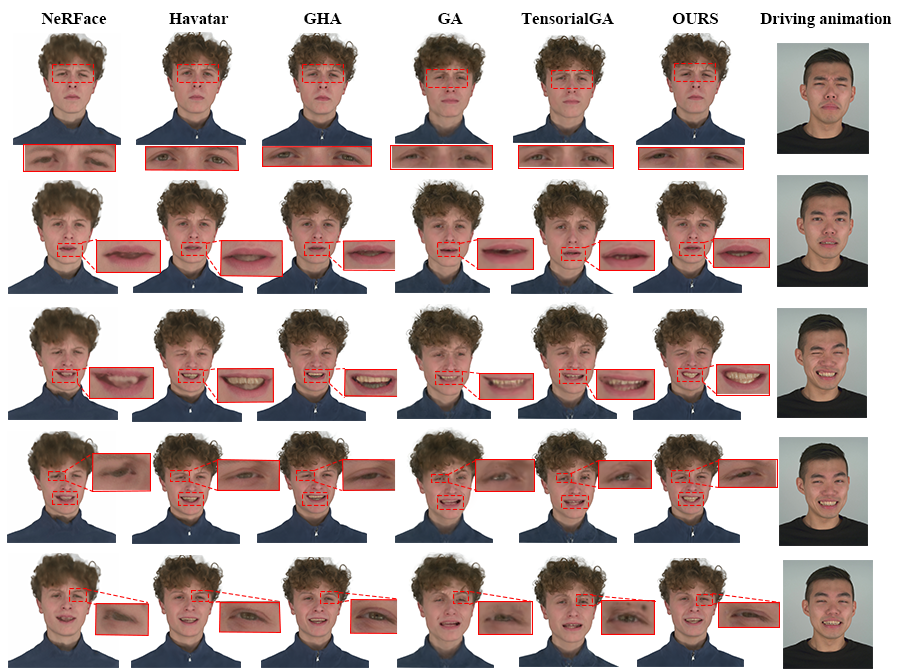}
  \caption{Qualitative comparisons of different methods on the cross-identity reenactment task. Each column represents a different method, and each row shows results at consecutive time steps in the reenactment sequence. From left to right: NeRFace, HAvatar, GHA, GaussianAvatars (GA), TensorialGA, and Ours. Our method more faithfully transfers expressions and maintains identity across frames, producing photo-realistic results.}
  \label{fig:cross_id}
\end{figure*}

In summary, while these methods utilize 3DGS to improve facial avatar reconstruction, they frequently encounter difficulties in accurately modeling fine-grained expression in regions of high deformation, thereby constraining their capacity to capture intricate and subtle facial dynamics. To address the challenges, we propose a novel 3D head avatar modeling approach that integrates both expression-aware and deformation-aware control mechanisms. Our method is designed for multi-view training, as the proposed framework relies on multi-view observations to obtain reliable geometric cues and structure-aware initialization. Specifically, we introduce a controllable Gaussian mechanism that identifies key Gaussians exhibiting substantial expression-induced deformation by applying an experimentally determined threshold to the predicted displacement. Subsequently, a spatial propagation strategy is employed to adjust neighboring Gaussians, facilitating more precise and localized control in expressive regions, such as the mouth and eyebrows. As a complementary refinement strategy, we further adopt a deformation-aware Gaussian splitting scheme that increases Gaussian density in highly deformable regions based on predicted deformation offsets, so as to better preserve local geometric detail. Additionally, to mitigate the geometry instability observed during the early training stages of existing methods, we introduce a generative prior derived from a pretrained large model. By designing a mesh alignment and structural supervision mechanism, this prior continuously guides the optimization of the geometry throughout training, significantly improving the reconstruction quality of head contours and occluded regions. Through the above components, our framework achieves high-fidelity expression reenactment across identities, as illustrated in Fig.~\ref{fig:cross_id} and Fig.~\ref{fig:cross_id_2}.

The main contributions of this work are summarized below.
\begin{itemize}
    \item We propose an expression-aware 3D head avatar reconstruction framework with a controllable Gaussian mechanism for localized deformation modeling. In addition, we incorporate a deformation-aware splitting refinement to improve local detail representation in highly deformable regions.

    \item We introduce a structure-aware geometry modeling module guided by generative priors from a large-scale generative model, which improves early-stage training stability and ensures globally consistent geometry.

    \item Our method is evaluated on multiple expression-driven benchmarks. The results demonstrate superior performance in terms of expression reconstruction accuracy, detail preservation, and identity consistency, showing strong generalization and practical value. 
\end{itemize}

\section{Related Work}
\textbf{Implicit Neural Head Avatars.}
Early controllable head avatar methods are largely built upon parametric face models and neural implicit representations. Traditional models such as 3DMM and FLAME~\cite{li2017learning} provide compact and interpretable control over facial shape, pose, and expression, but they are often limited in representing complex geometry and non-linear local deformations. With the development of neural rendering, implicit field-based methods such as NeRF~\cite{mildenhall2020nerf} and NeuS~\cite{wang2021neus} have substantially improved photorealism and personalization for head avatar reconstruction. Representative works in this line improve head avatar modeling in terms of controllability, efficiency, and synthesis fidelity. FaceNeRF~\cite{gafni2021facenerf} and NeRFace~\cite{gafni2021nerface} model dynamic facial appearance using expression-conditioned neural fields, while IMAvatar~\cite{zheng2022avatar} and Neural Head Avatars~\cite{grassal2022neural} further incorporate deformation constraints or structural guidance to improve controllability. More recent methods continue to strengthen this paradigm by improving local motion modeling, reconstruction speed, or rendering quality. For example, Chen et al.~\cite{chen2023implicit} introduce controllable local deformation fields to better capture localized and asymmetric facial motion, INSTA~\cite{zielonka2023instant} improves efficiency by combining dynamic neural radiance fields with geometry-guided neural graphics primitives around a parametric face model, MonoNPHM~\cite{giebenhain2024mononphm} formulates monocular dynamic head reconstruction using a neural parametric head model with coupled geometry and appearance representations, and DiffusionAvatars~\cite{kirschstein2024diffusionavatars} further enhances synthesis fidelity by combining NPHM-based proxy geometry with diffusion-based neural rendering. Overall, although these methods achieve strong reconstruction quality and controllability, they mainly rely on implicit fields, volumetric rendering, or diffusion-based image synthesis. In contrast, our method is built on an explicit dynamic Gaussian representation and focuses on local deformation modeling in highly expressive regions.

\textbf{Gaussian Splatting-based Head Avatars.}
Recently, 3D Gaussian Splatting has emerged as a promising alternative for head avatar modeling, owing to its favorable balance between rendering fidelity and efficiency. Existing Gaussian-based methods can be broadly grouped according to how they model dynamics and achieve controllability. One line focuses on deformable or rigged Gaussian representations for animation. For instance, GaussianHead~\cite{wang2025gaussianhead} models dynamic heads using deformable 3D Gaussians and a compact tri-plane with learnable derivations, D3GA~\cite{zielonka2025drivable} embeds 3D Gaussian primitives into tetrahedral cages to construct drivable avatars, where the facial component supports dynamic head motion, GHA~\cite{xu2024gaussian} improves reconstruction fidelity through dynamic Gaussians with expression-aware decomposition, and GaussianAvatars~\cite{qian2024gaussianavatars} rigs 3D Gaussians to a parametric face model to enable pose-, expression-, and viewpoint-controllable animation. Another line emphasizes more structured or compact Gaussian modeling. NPGA~\cite{giebenhain2024npga} introduces a data-driven parametric Gaussian formulation for high-fidelity controllable avatars, while Gaussian Eigen Models (GEM)~\cite{zielonka2025gaussian} compress dynamic avatars into compact low-dimensional spaces to improve lightweight controllability. A third line focuses more explicitly on local detail synthesis. ScaffoldAvatar~\cite{aneja2025scaffoldavatar} improves close-up realism by coupling patch-level expression modeling with hierarchical Gaussian generation, and HumanGaussian~\cite{zhang2024humangaussian} demonstrates monocular Gaussian reconstruction from a single image. Taken together, these methods highlight the strong potential of Gaussian-based representations for high-quality head animation. However, prior methods mainly focus on rigging mechanisms, parametric Gaussian modeling, lightweight controllability, or patch-level expression modeling. In contrast, our method addresses a complementary problem: improving the modeling of highly deformable facial regions by explicitly selecting strongly displaced Gaussians as control anchors, propagating their motion to neighboring Gaussians, and adaptively increasing Gaussian density in regions with large deformation.

\textbf{Neural Head Modeling with Structural Priors.}
Another related line introduces stronger structural priors into neural head modeling. Methods in this line mainly use semantic, parametric, or mesh-based priors to regularize geometry, reduce ambiguity in under-constrained regions, and improve controllability. To bridge quality and control, recent efforts have integrated 3DMM priors into neural frameworks~\cite{deng2019accurate,feng2021learning,gecer2019ganfit,romdhani2005estimating}. Semantic facial NeRF models, such as the personalized semantic facial NeRF of Gao et al.~\cite{gao2022reconstructing}, show that interpretable and disentangled structural guidance is important for controllability and stable reconstruction. IMAvatar~\cite{zheng2022avatar} blends shape bases and skinning fields for expression and pose deformation, but is dependent on tracking and iterative ray marching. i3DMM~\cite{yenamandra2021i3dmm} decouples shape into a reference geometry and deformation field, enabling dense correspondence but introducing noise in hair regions. Neural Head Avatars~\cite{grassal2022neural} separate geometry and texture via color-aware energy terms to improve local consistency under large expression or pose changes. While these priors are effective, in many existing methods they are mainly used for deformation parameterization or coarse geometric initialization. In contrast, our method introduces a generative geometry prior into a dedicated first-stage structure-aware geometry modeling process. By aligning the predicted geometry with a high-quality identity-specific prior mesh and then using the resulting surface to initialize the Gaussian representation, our method provides a more stable starting point for subsequent dynamic optimization. This design is particularly beneficial for preserving global geometric consistency while supporting fine-grained expression reconstruction in later stages.

In summary, recent methods have substantially advanced reconstruction fidelity, controllability, and efficiency for neural head avatars. Nevertheless, accurately modeling highly deformable facial regions while maintaining stable geometry optimization remains challenging. This motivates our expression-aware Gaussian control mechanism, deformation-aware splitting strategy, and prior-guided structure-aware geometry modeling.

\section{Proposed Method}
\begin{figure*}
\centering
  \includegraphics[width=\textwidth]{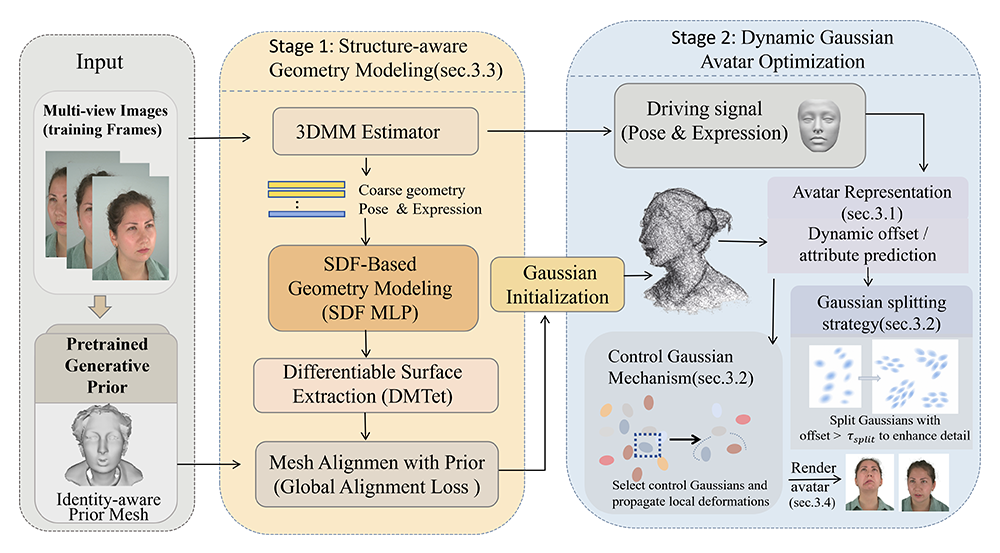}
  \caption{Overview of the proposed two-stage framework. In Stage~1 (Sec.~3.3), multi-view images are first processed by a 3DMM estimator to provide coarse geometry, pose, and expression cues. Based on these cues, the framework performs SDF-based geometry modeling, extracts a surface through differentiable surface extraction, aligns the resulting mesh with an identity-aware prior mesh, and then initializes the dynamic Gaussian avatar. In Stage~2 (Sec.~3.1, Sec.~3.2, and Sec.~3.4), the initialized avatar is driven by pose and expression signals to predict dynamic offsets and attribute variations, which are further refined in locally highly deformable regions through the proposed control Gaussian mechanism and Gaussian splitting, followed by differentiable rendering and reconstruction optimization to produce the final dynamic head avatar.}
  \Description{}
  \label{fig:network}
\end{figure*}

We aim to reconstruct a high-quality and expression-controllable 3D head avatar under large deformations. Our method is built upon a dynamic 3D Gaussian representation and adopts a two-stage framework, as illustrated in Fig.~\ref{fig:network}. Overall, the pipeline consists of three parts: the input data, Stage~1 for structure-aware geometry modeling, and Stage~2 for dynamic Gaussian avatar optimization.

In Section~\ref{sec:mesh_init}, we describe the first stage of training, which learns a structure-aware geometry from multi-view observations with the guidance of an identity-aware prior mesh derived from a generative model. Specifically, multi-view images are first processed by a 3DMM estimator to provide coarse geometry, pose, and expression cues, which are then used for SDF-based geometry modeling, differentiable surface extraction, and prior-guided mesh alignment. The resulting geometry is further converted into an initialized dynamic Gaussian avatar through Gaussian initialization.

Starting from this initialized avatar, the second stage performs dynamic refinement under pose and expression signals. In Section~\ref{sec:representation}, we introduce the basic modeling pipeline, which predicts dynamic offsets and attribute variations based on expression and pose parameters. Section~\ref{sec:control_gaussian} presents the proposed controllable Gaussian mechanism, which selects Gaussians with large expression-induced deformation and propagates local deformations to their neighbors for more stable local modeling. We further incorporate a deformation-aware Gaussian splitting refinement as an auxiliary design to enhance local detail representation in highly deformable regions. Finally, Section~\ref{sec:training} describes the overall optimization process, where differentiable rendering and reconstruction constraints are used to jointly refine appearance and expression control for the final dynamic head avatar.

\subsection{Avatar Representation}
\label{sec:representation}
Building on the theoretical principles of 3D Gaussian Splatting~\cite{kerbl20233d}, we have designed a dynamic avatar representation based on 3D Gaussian primitives capable of modelling geometric shapes as well as appearance changes driven by expressions and poses. Our approach uses a canonical Gaussian set as the base shape and applies learnable expression-aware transformations to generate personalized avatar representations.

We begin by fitting a 3D Morphable Model (3DMM) to each frame of a multi-view sequence, extracting two types of control parameters: expression coefficients $\theta \in \mathbb{R}^{d_{\text{exp}}}$ and head pose parameters $\beta \in \mathbb{R}^6$. Based on the canonical head pose, we construct a neutral Gaussian set:
\begin{equation}
\mathcal{G}_0 = \{ \mathbf{X}_0, \mathbf{F}_0, \mathbf{Q}_0, \mathbf{S}_0, \mathbf{A}_0 \},
\label{eq:neutral_gaussian}
\end{equation}
where $\mathbf{X}_0 \in \mathbb{R}^{N \times 3}$ are the positions of $N$ Gaussians, $\mathbf{F}_0$ denotes feature vectors, $\mathbf{Q}_0$ the rotations, $\mathbf{S}_0$ the scales, and $\mathbf{A}_0$ the opacities. Dynamic colors are predicted from $\mathbf{F}_0$ through learnable networks, to the extent that neutral colors do not have to be defined.

To model the deformation caused by $\theta$ and $\beta$, we introduce multiple attribute-specific MLPs for dynamic prediction. Each MLP predicts residuals to update attributes from the neutral template. Specifically, the spatial position is defined as:
\begin{equation}
\mathbf{X}(\theta, \beta) = \mathbf{X}_0 + f_{\text{exp}}^{\text{def}}(\mathbf{X}_0, \theta) + f_{\text{pose}}^{\text{def}}(\mathbf{X}_0, \beta).
\label{eq:position_update}
\end{equation}

On the other hand, color attributes are dynamically predicted by:
\begin{equation}
\mathbf{C}(\theta, \beta) = f_{\text{exp}}^{\text{color}}(\mathbf{F}_0, \theta) + f_{\text{pose}}^{\text{color}}(\mathbf{F}_0, \beta).
\label{eq:color_update}
\end{equation}

Other properties, including rotation $Q$, scale $S$, and opacity $A$, follow the same residual modeling strategy:
\begin{align}
\mathbf{Q}(\theta, \beta) &= \mathbf{Q}_0 + f_{\text{exp}}^{\text{rot}}(\mathbf{Q}_0, \theta) + f_{\text{pose}}^{\text{rot}}(\mathbf{Q}_0, \beta),
\label{eq:rot_update} \\
\mathbf{S}(\theta, \beta) &= \mathbf{S}_0 + f_{\text{exp}}^{\text{scale}}(\mathbf{S}_0, \theta) + f_{\text{pose}}^{\text{scale}}(\mathbf{S}_0, \beta), \label{eq:scale_update} \\
\mathbf{A}(\theta, \beta) &= \mathbf{A}_0 + f_{\text{exp}}^{\text{alpha}}(\mathbf{A}_0, \theta) + f_{\text{pose}}^{\text{alpha}}(\mathbf{A}_0, \beta). \label{eq:alpha_update}
\end{align}

To transform all spatial attributes into the world coordinate system, we apply a rigid transform $T(\cdot)$ to the positions and rotations:
\begin{equation}
\{ \mathbf{X}^{\text{world}}, \mathbf{Q}^{\text{world}} \} = T( \{ \mathbf{X}(\theta, \beta), \mathbf{Q}(\theta, \beta) \} ).
\label{eq:rigid_transform}
\end{equation}

The remaining attributes $\mathbf{C}, \mathbf{S}, \mathbf{A}$ are preserved in the local coordinate space. The final expression-aware Gaussian set is defined as:
\begin{equation}
\mathcal{G}(\theta, \beta) = \left\{ \mathbf{X}^{\text{world}}, \mathbf{C}, \mathbf{Q}^{\text{world}}, \mathbf{S}, \mathbf{A} \right\}.
\label{eq:final_gaussian}
\end{equation}

This set $\mathcal{G}(\theta, \beta)$ (Eq.~\ref{eq:final_gaussian}) is passed to the differentiable renderer for image synthesis, enabling expression-driven reconstruction and high-fidelity avatar rendering. Unlike prior works that treat all Gaussians uniformly, our full pipeline introduces a control-based selection mechanism and Gaussian splitting strategy (see Sec.~\ref{sec:control_gaussian}) to enhance local expressiveness and handle high-deformation regions more effectively.

\subsection{Controllable Gaussian Mechanism}
\label{sec:control_gaussian}
In real facial expressions, particularly during intense motions such as laughing or frowning, key expressive areas (e.g., the mouth corners or brows) undergo significant deformations, while surrounding areas exhibit subtle changes in texture and geometry. Relying solely on global MLP-based prediction often fails to capture these local variations, resulting in blurry reconstructions or geometric discontinuities.

To address this, we propose a controllable Gaussian mechanism that enhances the expressiveness of highly deformable areas by explicitly modeling local geometric adjustments. Specifically, we automatically identify \textit{Control Gaussians} based on the magnitude of predicted displacements and propagate their influence to neighboring Gaussians via distance-weighted interpolation. This allows our model to accurately capture localized deformations while preserving consistency in high-expression regions.

Specifically, we start by computing the expression-driven displacement magnitude of each Gaussian using the MLP $f_{\text{exp}}^{\text{def}}$ from the base model:
\begin{equation}
\Delta x_i = \left\| f_{\text{exp}}^{\text{def}}(\mathbf{x}_i, \theta) \right\|_2,
\label{eq:displacement_magnitude}
\end{equation}
where $\mathbf{x}_i \in \mathbb{R}^3$ is the canonical position of the $i$-th Gaussian, and $f_{\text{exp}}^{\text{def}}(\mathbf{x}_i, \theta)$ outputs its predicted expression offset. We then define a threshold $\tau$ to detect significantly displaced Gaussians. If $\Delta x_i > \tau$, we classify the $i$-th Gaussian as a \textit{Control Gaussian}. The set of Control Gaussians is denoted as:
\begin{equation}
\mathcal{C} = \left\{ i \mid \Delta x_i > \tau, \ i = 1, \dots, N \right\}.
\label{eq:control_gaussian_set}
\end{equation}

To enhance local consistency, we propagate the influence of each Control Gaussian to its nearby neighbors. For each $\mathbf{x}_i \in \mathcal{C}$, we use nearest neighbor search to find a local neighborhood:
\begin{equation}
\mathcal{N}(i) = \left\{ j \mid \left\| \mathbf{x}_j - \mathbf{x}_i \right\|_2 < r, \ j \ne i \right\},
\label{eq:neighbor_set}
\end{equation}
where $r$ is a predefined radius controlling the influence range. Notably, a neighboring point $x_j$ may be influenced by multiple control points. To model this multi-source influence, we define a set of control Gaussians $C_j$ that affect each $x_j$. Here, $x'_i = x_i + f^{\mathrm{def}}_{\mathrm{exp}}(x_i,\theta)$ denotes the expression-updated position of the $i$-th Control Gaussian, and thus $(x'_i - x_i)$ is the corresponding deformation vector, whose magnitude is used in Eq.~(9). The final position of $x_j$ is then updated as:
\begin{equation}
x'_j = x_j + \sum_{i \in C_j} w_{ij}(x'_i - x_i), 
\label{eq:multi_control_update}
\end{equation}
where the influence weight $w_{ij}$ is computed based on the spatial distance between $x_j$ and each control point $x_i$ using a Gaussian kernel, normalized within $C_j$:
\begin{equation}
w_{ij} = \frac{
\exp\left( -\frac{ \|\mathbf{x}_j - \mathbf{x}_i\|_2^2 }{ \sigma^2 } \right)
}{
\sum_{k \in \mathcal{C}_j} \exp\left( -\frac{ \|\mathbf{x}_j - \mathbf{x}_k\|_2^2 }{ \sigma^2 } \right)
}.
\label{eq:gaussian_weight}
\end{equation}

Here, $\sigma$ controls the decay of influence with respect to spatial distance. In practice, this interaction is restricted to sparse local neighborhoods, so that only selected Control Gaussians participate in the aggregation rather than dense global all-to-all interaction. This neighborhood-aware adjustment---realized through Gaussian-weighted aggregation from multiple control points---promotes smoother local deformations while preserving sharp details in highly expressive regions, as reflected in Eq.~\ref{eq:multi_control_update} and Eq.~\ref{eq:gaussian_weight}. Through this mechanism, our method enhances the representation of fine-grained dynamics and improves realism in highly deformable areas, especially under extreme expressions.

Although a 3DMM-based prior can provide coarse guidance for identifying highly deformable facial regions, it may be less flexible for capturing subject-specific and highly localized motions. In contrast, our method selects Control Gaussians directly from the predicted displacement magnitude, making the selection adaptive to the actual deformation patterns learned for each subject and expression.

\textit{Gaussian Splitting Strategy.} In expressive regions with extremely large deformations, a single Gaussian may struggle to accurately model complex local geometry. To address this issue, we introduce a targeted Gaussian splitting strategy. When the predicted displacement of a Gaussian exceeds a higher threshold $\tau_{\text{split}}$, we dynamically split it into two split instances. The split instances are initialized close to the original location and inherit its appearance and structural attributes. This process increases the local density of Gaussians in highly deformable regions, enabling a finer representation of geometric variations. Our splitting strategy is deformation-aware and only applies to regions with large displacements. This ensures modeling efficiency while effectively improving local detail preservation and geometric expressiveness under extreme facial motions.

\subsection{Structure-Aware Geometry Modeling with Generative Prior Constraints}
\label{sec:mesh_init}
To provide a stable and structure-aware foundation for downstream surface representation, we introduce a geometry modeling stage guided by a high-quality prior mesh generated by a large-scale generative model. This first-stage module optimizes an implicit signed distance function (SDF) via a neural network, from which a differentiable mesh surface is extracted using DMTet. The resulting mesh serves as reliable and identity-aware guidance for Gaussian initialization in the subsequent stage.

We first represent the geometry using an implicit signed distance function (SDF) modeled by a multi-layer perceptron (MLP) $f_{\text{sdf}}$, which maps a 3D point $\mathbf{x} \in \mathbb{R}^3$ to a scalar SDF value $s$ and a feature vector $\eta$:
\begin{equation}
(s, \eta) = f_{\text{sdf}}(\mathbf{x}).
\label{eq:sdf}
\end{equation}
Here, $s \in \mathbb{R}$ denotes the signed distance to the surface (positive for outside, negative for inside), and $\eta \in \mathbb{R}^d$ encodes features used for downstream appearance prediction. We then extract the initial implicit surface using Deep Marching Tetrahedra (DMTet)~\cite{shen2021deep}, as it supports differentiable surface extraction for end-to-end training:
\begin{equation}
\hat{\mathbf{X}} = \mathrm{DMTet}(f_{\text{sdf}}),
\label{eq:dmtet}
\end{equation}
where $\hat{\mathbf{X}} \in \mathbb{R}^{M \times 3}$ is the resulting set of $M$ mesh vertices. To improve the robustness and structural accuracy of the extracted mesh in the training stage, we incorporate a high-quality prior mesh as additional geometric guidance. We denote the prior mesh as:
\begin{equation}
\mathbf{X}^{\text{mesh}} = \{ \mathbf{x}_m^{\text{mesh}} \}_{m=1}^{M}, \quad \mathbf{x}_m^{\text{mesh}} \in \mathbb{R}^3
\label{eq:prior_mesh}
\end{equation}
where each $\mathbf{x}_m^{\text{mesh}}$ represents the 3D position of the $m$-th vertex. We generate this 3D mesh using a large-scale pretrained generative model~\cite{xiang2024structured}, which adopts a structured latent representation and a sparse-aware transformer architecture, enabling more accurate and identity-specific initialization shapes.

Subsequently, we apply Iterative Closest Point (ICP)~\cite{besl1992method} to align the prior mesh to our predicted mesh, ensuring coordinate consistency. After alignment, we treat the prior mesh as a structural constraint to guide the optimization toward more accurate target shapes.  
Specifically, we introduce a global alignment loss that enforces consistency in the overall shape and scale of the predicted mesh with respect to the prior.  
Rather than relying on point-wise vertex supervision—which tends to be noisy and overly restrictive at this stage—we design a lightweight yet effective constraint based on two stable geometric features: the mesh center and scale. We compute each mesh's global center as the mean of its vertices, and define the scale as the mean distance from the vertices to the center. Given the prior mesh (pre-aligned via ICP) and the predicted mesh, we denote their centers as $\mathbf{c}^{\text{mesh}}$ and $\hat{\mathbf{c}}$, and their scales as $s^{\text{mesh}}$ and $\hat{s}$, respectively. The alignment loss is defined as:
\begin{equation}
\mathcal{L}_{\text{mesh}} = \| \mathbf{c}^{\text{mesh}} - \hat{\mathbf{c}} \|_2^2 + (s^{\text{mesh}} - \hat{s})^2.
\label{eq:loss_mesh}
\end{equation}

This prior-guided constraint ensures global geometric consistency while avoiding overly strict local supervision, thereby enhancing the stability and structural integrity of the predicted mesh. Unlike a fixed facial template, our prior mesh is generated from the input image using a powerful generative model, enabling identity- and shape-specific adaptation. This adaptation leads to more accurate geometry modeling and stronger structural alignment during optimization. As shown in Fig.~\ref{fig:init_ablation}, our method yields more plausible and consistent geometry when trained for the same number of epochs in the first stage, demonstrating improved convergence behavior. Our ablation studies (see Fig.~\ref{fig:init_ablation} and Table~\ref{tab:ablation-results}) confirm that removing this structure-aware modeling strategy leads to inferior surface reconstruction and unstable optimization, demonstrating its critical importance to the overall framework.

In summary, the proposed structure-aware geometry modeling module, guided by a generative prior and optimized through global alignment, significantly enhances geometric accuracy and consistency. It establishes a stable structural foundation that facilitates high-fidelity expression reconstruction in subsequent stages.

\subsection{Training}
\label{sec:training}
We employ a two-stage training strategy to ensure stable convergence and high-quality avatar reconstruction.

\textbf{Stage I: Structural Geometry Modeling.}  
In the first stage, we jointly optimize the expression- and pose-driven deformation networks along with the neutral Gaussian set and the implicit surface. The objective integrates several commonly used supervision signals, including RGB reconstruction, silhouette consistency, landmark proximity, Laplacian regularization, and deformation offset regularization. Importantly, we introduce a novel global alignment loss based on the global center and scale of the predicted and prior meshes. As presented in Sec.~\ref{sec:mesh_init}, this large-scale prior mesh alignment loss constrains the predicted geometry using a high-quality mesh generated by a large-scale pretrained model. This global constraint is robust to local noise and significantly improves early-stage geometric stability. The total loss for this stage is defined as:
\begin{align}
\mathcal{L}_{\text{init}} = 
& \ \lambda_{\text{rgb}} \mathcal{L}_{\text{rgb}} + \lambda_{\text{sil}} \mathcal{L}_{\text{sil}} + 
\lambda_{\text{offset}} \mathcal{L}_{\text{offset}} \nonumber \\
& + \lambda_{\text{lmk}} \mathcal{L}_{\text{lmk}} + 
\lambda_{\text{lap}} \mathcal{L}_{\text{lap}} + 
\lambda_{\text{mesh}} \mathcal{L}_{\text{mesh}}.
\label{eq:init}
\end{align}

To balance the contributions of different supervision signals during training, we empirically set the loss weights as follows: $\lambda_{\text{rgb}} = 1.0$, $\lambda_{\text{sil}} = 0.1$, $\lambda_{\text{offset}} = 0.01$, $\lambda_{\text{lmk}} = 0.1$, $\lambda_{\text{lap}} = 100$, and $\lambda_{\text{mesh}} = 1.0$. These values are kept fixed throughout all experiments to ensure consistency and fair comparison across different model variants.

\textbf{Stage II: Dynamic Optimization.}  
In the second stage, starting from the stabilized geometry obtained in Stage I, we continue to jointly optimize the expression, pose, and attribute networks along with all other components in an end-to-end manner. This stage further refines the dynamic Gaussian representation in terms of geometry, appearance, and controllability. Rather than continuing the explicit geometric regularization used in Stage I, Stage II focuses on dynamic refinement under image-space supervision. We adopt a combination of RGB reconstruction loss and LPIPS perceptual loss on randomly cropped visible local patches:
\begin{equation}
\mathcal{L}_{\text{total}} = 
\lambda_{\text{rgb}} \mathcal{L}_{\text{rgb}} + 
\lambda_{\text{lpips}} \mathcal{L}_{\text{lpips}}.
\label{eq:total}
\end{equation}
Here, $L_{rgb}$ denotes the overall RGB reconstruction term. In practice, the RGB supervision is applied at both coarse and high-resolution scales, while $L_{lpips}$ is computed on randomly cropped visible local regions. This design provides sufficient observation-level constraints for stable optimization, while preserving the flexibility needed to model expression-dependent local deformations. We empirically set $\lambda_{rgb}=1.0$ and $\lambda_{lpips}=0.1$. This two-stage pipeline enables a smooth transition from stable geometry initialization to expressive, high-fidelity avatar reconstruction.

\section{Experiments}
In this section, we present the details of our experimental setup, datasets, ablation studies, and comparisons with existing methods. We begin by describing the implementation details, including training configurations and hardware specifications. Next, we introduce the datasets used for training and evaluation. Our method is evaluated on two tasks: self-reenactment and cross-identity reenactment. We report both qualitative and quantitative results, and compare our approach against state-of-the-art methods to demonstrate its effectiveness. We also provide a more visual video comparison of the results in the project. Additionally, we conduct a series of ablation studies to investigate the impact of each key module in our framework. By incrementally adding the proposed modules, we analyze their individual contributions to the final performance through both visual and numerical comparisons.

\subsection{Datasets.} We conduct our experiments on multi-view video data from the NeRSemble dataset~\cite{kirschstein2023nersemble}, which contains 16 camera view sequences for each subject. Each camera captures multiple expression sequences, such as EMO-1-shout+laugh. For each identity, we use sequences labeled as “FREE” for evaluation and the remaining sequences for training.
Following the preprocessing pipeline of GHA~\cite{xu2024gaussian}, we first remove the background from each frame using a segmentation model~\cite{lin2021real}, and extract 68 facial 2D landmarks using a standard landmark detector~\cite{bulat2017far}. 
We then fit the multi-view Basel Face Model (BFM)~\cite{gerig2018morphable} to estimate 3D landmarks, expression coefficients, and head pose corresponding to the detected 2D landmarks.

\subsection{Evaluation Metrics.} For quantitative evaluation, we adopt three widely used metrics: Peak Signal-to-Noise Ratio (PSNR), Structural Similarity Index (SSIM), and Learned Perceptual Image Patch Similarity (LPIPS). These metrics assess the reconstruction quality from the perspectives of pixel-level fidelity, structural consistency, and perceptual similarity, respectively.

\subsection{Training Details.} We use the Adam optimizer for both training stages. In the structural geometry modeling stage, the learning rate is set to $1 \times 10^{-3}$ for all networks. We train this stage for 10{,}000 iterations with a batch size of 4. In the second stage, we jointly optimize all components of the Gaussian avatar model. The learning rates are set to $1 \times 10^{-4}$ for the color, deformation, and attribute MLPs; $1 \times 10^{-5}$ for the neutral positions $\mathbf{X}_0$ and feature vectors $\mathbf{F}_0$; $1 \times 10^{-5}$ for rotation $\mathbf{Q}_0$; $3 \times 10^{-5}$ for scale $\mathbf{S}_0$. This stage is trained for 500{,}000 iterations with a batch size of 1. It is worth noting that using the generative prior from a large-scale pretrained model to guide structural geometry modeling reduces the overall training time by 11\% compared to not using the prior. We empirically set the control and splitting thresholds to 0.3 and 0.2, respectively, based on the typical size and movement range of expressive facial regions. The higher control threshold focuses on clearly deformed areas, while the lower splitting threshold enables earlier refinement in moderately changing regions. We will further discuss the reasonableness of these threshold choices in Sec.~\ref{sec:threshold}

\begin{figure*}
\centering
  \includegraphics[width=\textwidth]{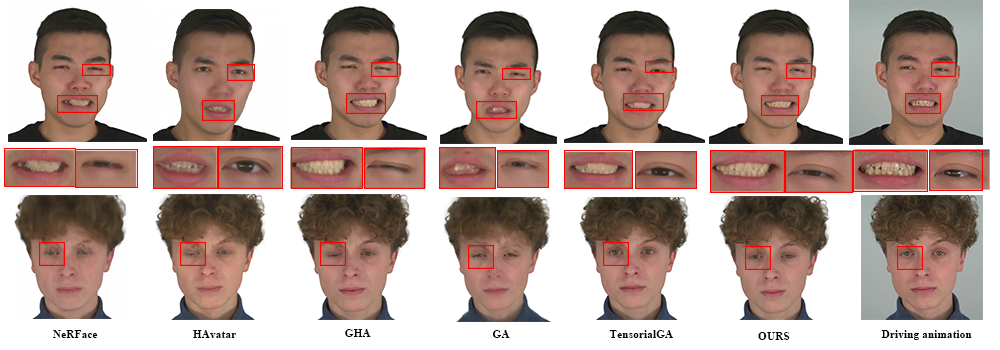}
  \caption{Qualitative comparisons of different methods on self-reenactment task. From left to right: NeRFace, HAvatar, GHA, GaussianAvatars(GA), TensorialGA and Ours. Our method can reconstruct details like eyes, teeth, etc. with high quality.}
  \label{fig:self_reenactment}
\end{figure*}

\subsection{Results and Comparisons}
For all compared methods, we follow the training settings reported in their original papers as closely as possible. Compared with the original data split used in GaussianAvatars and TensorialGA, our evaluation protocol is more challenging because it uses the FREE sequence for testing under a unified setting. During training, we observed that under this setting GaussianAvatars and TensorialGA were not sufficiently converged after 600k iterations and still produced noticeably blurry renderings. Therefore, we extended the training of these two methods to 800k iterations. The remaining compared methods follow the training schedules reported in their original papers.

\textbf{Self-Reenactment Evaluation.} We qualitatively and quantitatively compare our method with several representative approaches on the self-reenactment task. Specifically, NeRFace~\cite{gafni2021nerface} employs a deep MLP to model an expression-conditioned dynamic NeRF, where facial expressions and head poses derived from a 3DMM are used as conditioning inputs to enable controllable face geometry and appearance modeling in a canonical space. HAvatar~\cite{zhao2023havatar} takes a 3DMM mesh as conditional input and adopts tri-plane-based neural radiance fields for high-fidelity reconstruction. Gaussian Head Avatar (GHA)~\cite{xu2024gaussian} builds an explicit head representation with dynamic 3D Gaussians, enabling ultra-high-fidelity image synthesis. GaussianAvatars~\cite{qian2024gaussianavatars} rigs 3D Gaussian splats to a parametric face model, achieving photorealistic head avatars controllable in pose, expression, and viewpoint. TensorialGA~\cite{wang20253d} further introduces a compact tensorial representation for dynamic head avatars, where static appearance is encoded in tri-planes and dynamic texture details are modeled by lightweight feature lines for efficient and expressive rendering. For fair comparison, we replace the adversarial GAN loss in HAvatar with the VGG perceptual loss, following the setting of Gaussian Head Avatar~\cite{xu2024gaussian}. In addition, we use the same input sequences and render all methods under a unified rendering pipeline.

\begin{figure*}
\centering
  \includegraphics[width=\textwidth]{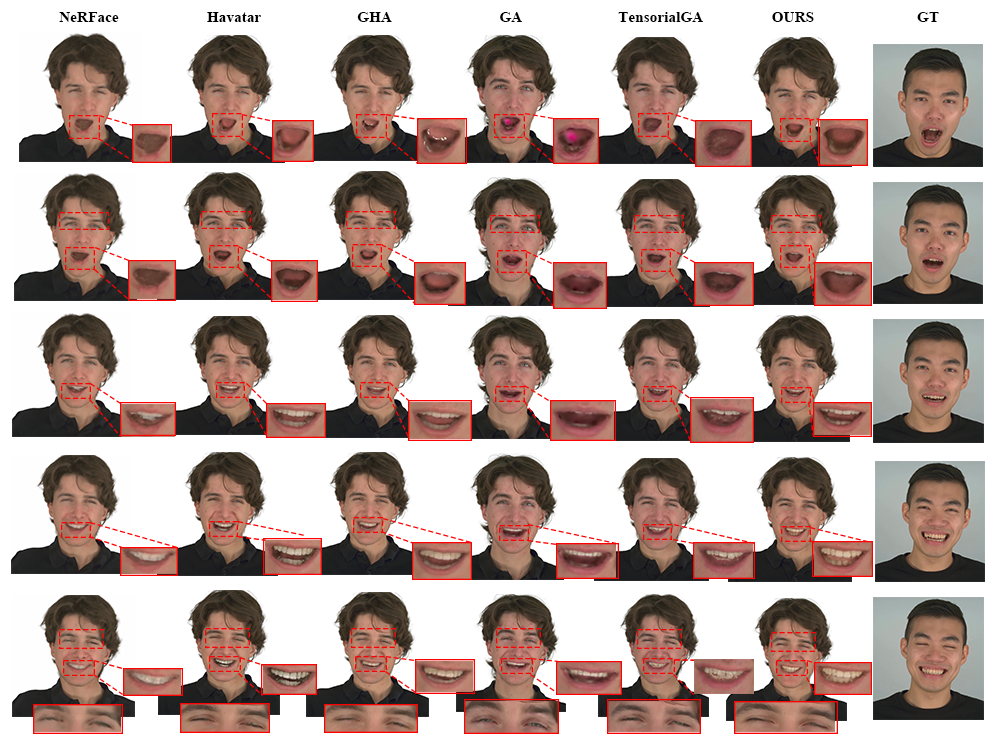}
  \caption{Another cross-identity example evaluated under the same setting as Fig.~\ref{fig:cross_id}. Columns correspond to methods and rows to consecutive frames. Our approach continues to exhibit consistent identity preservation and expression fidelity over time.}
  \label{fig:cross_id_2}
\end{figure*}

Qualitative comparisons are shown in Fig.~\ref{fig:self_reenactment}, where our method achieves sharper reconstruction and more faithful appearance in high-frequency regions such as teeth and hair. Furthermore, our model demonstrates better expression transferability than previous methods, for example in the accuracy of eye closure. More detailed video results can be found in our project.

\begin{table}[t]
  \centering
  \caption{Quantitative evaluation results of NeRFace, HAvatar, GHA, GaussianAvatars,  TensorialGA and our full method on the self-reenactment task. $\downarrow$ indicates lower is better, $\uparrow$ indicates higher is better.}
  \label{tab:results}
  \small

  \begin{tabular}{lccc|ccc|ccc}
    \toprule
    Method &
    \multicolumn{3}{c|}{Case 1} &
    \multicolumn{3}{c|}{Case 2} &
    \multicolumn{3}{c}{Case 3} \\
    \cmidrule(lr){2-4} \cmidrule(lr){5-7} \cmidrule(lr){8-10}
    & PSNR $\uparrow$ & SSIM $\uparrow$ & LPIPS $\downarrow$
    & PSNR $\uparrow$ & SSIM $\uparrow$ & LPIPS $\downarrow$
    & PSNR $\uparrow$ & SSIM $\uparrow$ & LPIPS $\downarrow$ \\
    \midrule
    NeRFace & 21.38 & 0.746 & 0.283 & 21.08 & 0.847 & 0.279 & 20.19 & 0.817 & 0.223 \\
    HAvatar  & 22.62 & 0.802 & 0.247 & 21.63 & 0.861 & 0.264 & 22.16 & 0.878 & 0.184 \\
    GHA      & 24.02 & 0.814 & 0.203 & 25.47 & 0.879 & 0.147 & 24.56 & 0.902 & 0.144 \\
    GaussianAvatars       & 21.61 & 0.796 & 0.237 & 25.73 & 0.850 & 0.151 & 21.23 & 0.846 & 0.197 \\
    TensorialGA &23.23 &0.807 & 0.213 & 24.29 & 0.860 & 0.242 & 23.60 & 0.895 & 0.162\\
    Ours     & \textbf{24.24} & \textbf{0.814} & \textbf{0.203}
             & \textbf{26.72} & \textbf{0.884} & \textbf{0.141}
             & \textbf{24.82} & \textbf{0.903} & \textbf{0.141} \\
    \bottomrule
  \end{tabular}

  \vspace{2mm}

  \begin{tabular}{lccc|ccc|ccc}
    \toprule
    Method &
    \multicolumn{3}{c|}{Case 4} &
    \multicolumn{3}{c|}{Case 5} &
    \multicolumn{3}{c}{Average} \\
    \cmidrule(lr){2-4} \cmidrule(lr){5-7} \cmidrule(lr){8-10}
    & PSNR $\uparrow$ & SSIM $\uparrow$ & LPIPS $\downarrow$
    & PSNR $\uparrow$ & SSIM $\uparrow$ & LPIPS $\downarrow$
    & PSNR $\uparrow$ & SSIM $\uparrow$ & LPIPS $\downarrow$ \\
    \midrule
    NeRFace & 20.58 & 0.825 & 0.219 & 21.04 & 0.793 & 0.239 & 20.85 & 0.805 & 0.248 \\
    HAvatar  & 21.70 & 0.839 & 0.215 & 21.54 & 0.801 & 0.231 & 21.93 & 0.836 & 0.228 \\
    GHA      & 25.49 & 0.859 & 0.164 & 23.28 & 0.812 & 0.217 & 24.56 & 0.853 & 0.175 \\
    GaussianAvatars    &21.99 & 0.856 & 0.194 & 21.96 & 0.782 & 0.258 & 22.50 & 0.826 & 0.207 \\  
    TensorialGA & 25.48 & 0.862 & 0.194 & 23.75 & 0.814 & 0.210 & 24.07 & 0.848 & 0.204 \\
    Ours     & \textbf{25.92} & \textbf{0.863} & \textbf{0.161}
             & \textbf{23.66} & \textbf{0.822} & \textbf{0.196}
             & \textbf{25.07} & \textbf{0.857} & \textbf{0.168} \\
    \bottomrule
  \end{tabular}
\end{table}

We also perform a quantitative comparison across five identities using three standard metrics: Peak Signal-to-Noise Ratio (PSNR), Structural Similarity Index (SSIM), and Learned Perceptual Image Patch Similarity (LPIPS). Unlike the previous method Gaussian Head Avatar~\cite{xu2024gaussian}, which selects six random camera views for evaluation, we compute the average metric over all 16 camera views to comprehensively assess multi-view performance. In addition, we retain the full head region—including the neck and shoulders—without using facial parsing to remove body areas, since these regions are also important for identity modeling. The evaluation results are summarized in Table~\ref{tab:results}. Our method shows slight improvements in LPIPS and SSIM, while achieving a significant boost in PSNR, indicating better preservation of high-frequency details in reconstruction.

\begin{figure*}
\centering
\includegraphics[width=0.8\textwidth]{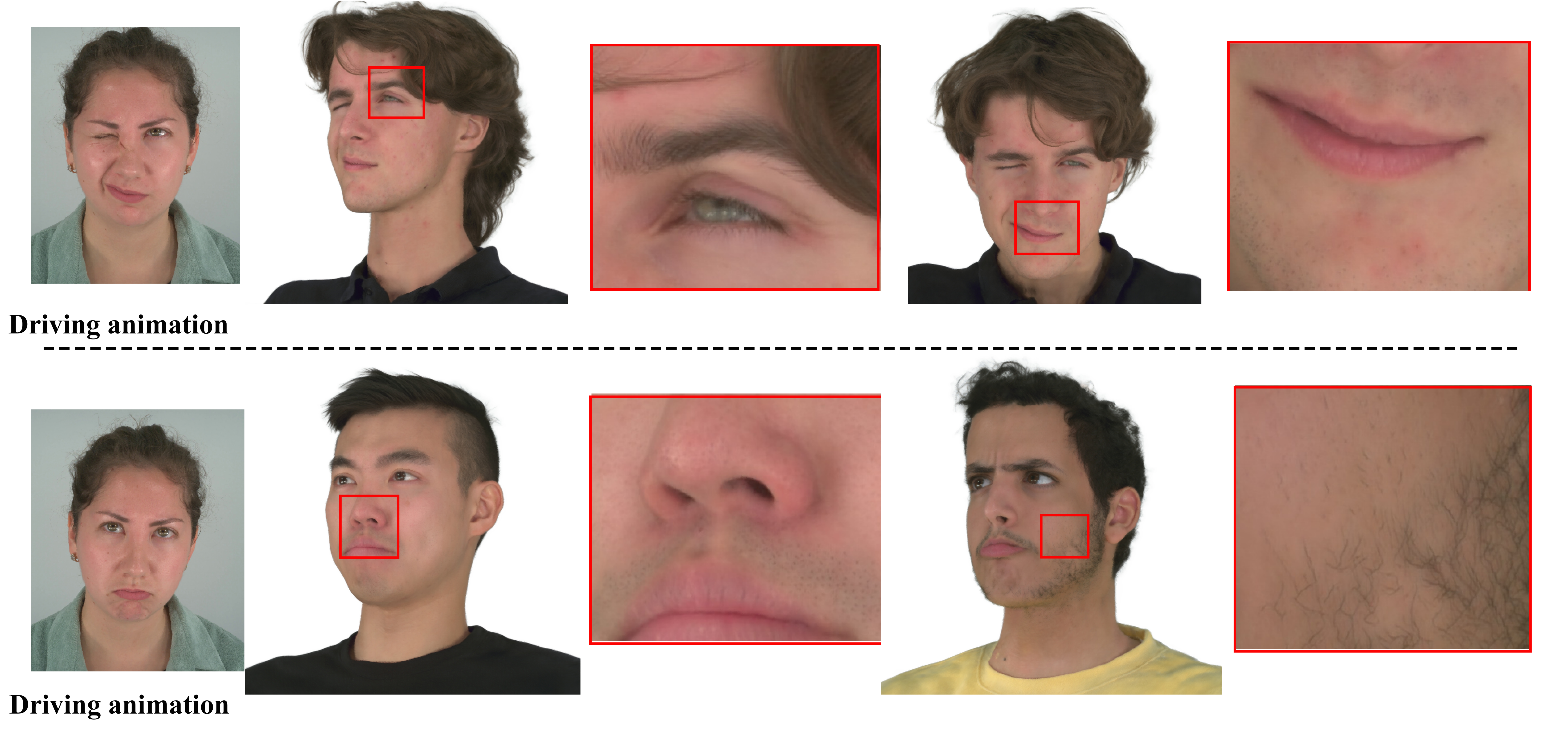}
  \caption{Results of our expression-controllable head avatar generation. As demonstrated, the generated results accurately capture facial expressions while preserving fine-grained local details.}
  \Description{}
  \label{fig:first_show}
\end{figure*}

\textbf{Cross-Identity Reenactment.} We also compare our method with previous leading methods on the cross-identity reenactment task.  Cross-identity reenactment is a generative task without ground-truth targets under novel expressions, so metrics cannot be computed. Many prior works rely on visual comparisons to evaluate expression transfer quality. To this end, we provide a qualitative comparison across identities. As shown in Fig.~\ref{fig:cross_id} and Fig.~\ref{fig:cross_id_2}, our approach generates clearer and more realistic results, with more accurate expression transfer. And as shown in Fig.~\ref{fig:first_show}, our method effectively captures facial expressions while preserving fine-grained local details.

\begin{table}[ht]
  \centering
  \caption{Quantitative evaluation of our method and other SOTA methods on 3D consistency for novel view synthesis.}
  \label{tab:novel_view}
  \small  %
  \begin{tabular}{lccc}
    \toprule
    Method & PSNR $\uparrow$ & SSIM $\uparrow$ & LPIPS $\downarrow$ \\
    \midrule
    NeRFace & 20.43 & 0.842 & 0.215 \\
    HAvatar & 21.16 & 0.878 & 0.184 \\
    GHA     & 22.89 & 0.880 & 0.167 \\
    GaussianAvatars  & 19.83 & 0.844 & 0.217 \\
    TensorialGA  & 21.51 & 0.876 & 0.182 \\
    Ours    & \textbf{23.22} & \textbf{0.882} & \textbf{0.165} \\
    \bottomrule
  \end{tabular}
\end{table}

\begin{figure*}
\centering
  \includegraphics[width=\textwidth]{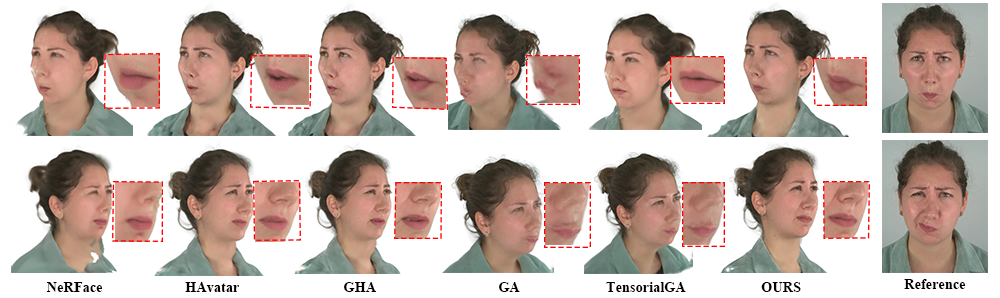}
  \caption{Qualitative comparison with prior methods on the novel view synthesis task. We use 8-view synchronized videos for training the avatar and the remaining 8 new views were used to test.}
  \Description{}
\label{fig:novel_view}
\end{figure*}

\textbf{Novel View Synthesis.} To evaluate the model’s ability to generate consistent images from new viewpoints, we conduct experiments on novel view synthesis. This task assesses the 3D consistency of the learned geometry and the model’s generalization across unseen camera angles. We train the model using data from only 8 camera views and test it on the remaining 8 unseen views. Fig~\ref{fig:novel_view} shows qualitative results under novel viewpoints. In addition, we perform a quantitative comparison with previous methods by computing PSNR, SSIM, and LPIPS on the test views. The average assessment results for the 8 test camera views are summarised in Table~\ref{tab:novel_view}. These results demonstrate that our method generalizes better to unseen views and maintains more consistent geometry across different camera angles.

\begin{figure}[h]
  \centering
  \includegraphics[width=0.7\linewidth]{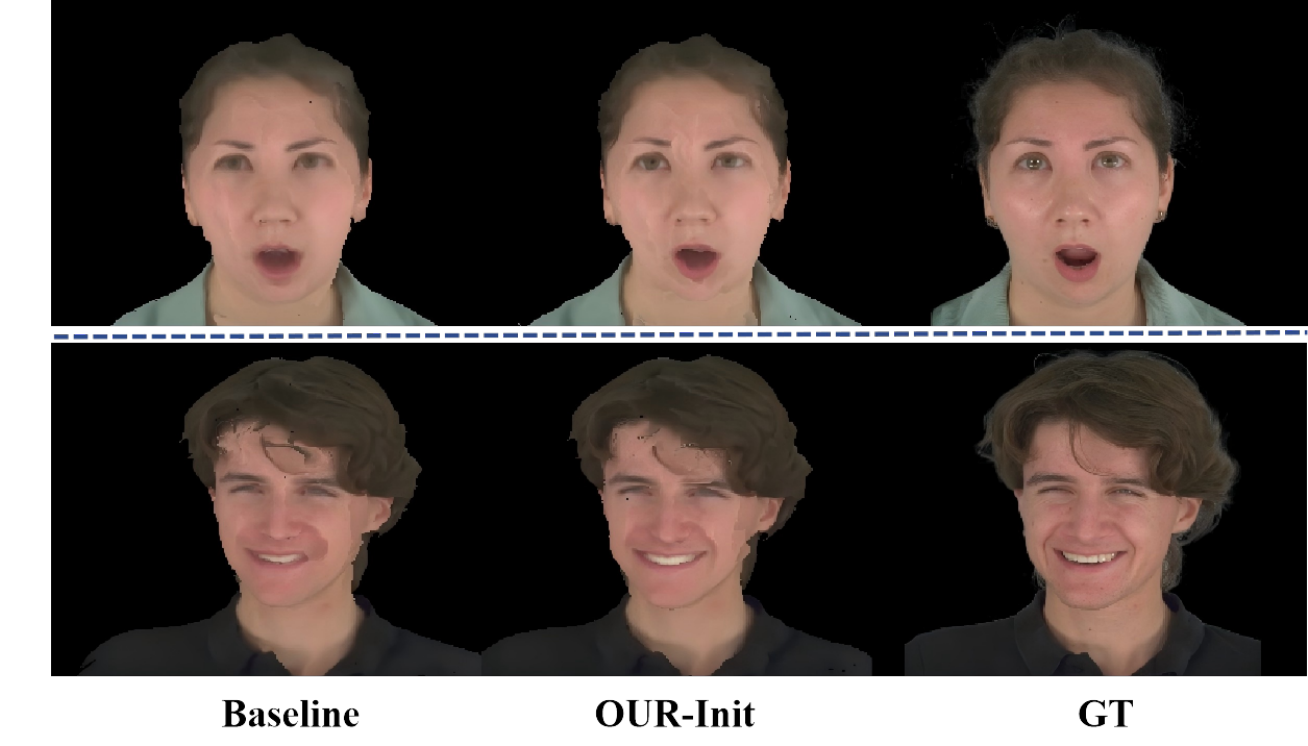}
  \caption{Ablation study of the structure-aware modeling strategy: by introducing an explicit mesh as a shape constraint, our strategy ensures that better contour and pentagonal shapes are obtained in the first stage.}
  \label{fig:init_ablation}
\end{figure}

\subsection{Ablation Study}
\subsubsection{Structure-aware modeling strategy.} To evaluate the effectiveness of our structure-aware modeling strategy, we conduct a comparison with a baseline that does not use any generative prior. In the baseline setup, the initial mesh is obtained by fitting an implicit SDF and color field without any geometric constraints. The resulting mesh vertices are directly used as Gaussian positions, and other attributes are optimized using multi-view image supervision. In contrast, our method adds a mesh-based constraint generated by a large-scale generative model. As shown in Fig~\ref{fig:init_ablation}, with the same epoch of training our method can achieve more stable and coherent modelling in regions such as contours and facial features. Moreover, by comparing the fourth and fifth rows in Table~\ref{tab:ablation-results}, we observe that incorporating our structure-aware modeling strategy (Ours, fifth row) leads to significant improvements across all reconstruction metrics. These results confirm that the proposed strategy improves the overall mesh quality and provides a stronger starting point for subsequent optimization.

\begin{table}[h]
  \centering
  \caption{Impact of the different modules proposed in our approach on the self-reenactment task.}
  \label{tab:ablation-results}
  \small
  \begin{tabular}{lccc|ccc}
    \toprule
    Method & \multicolumn{3}{c|}{Case 1} & \multicolumn{3}{c}{Case 2} \\
    \cmidrule(lr){2-4} \cmidrule(lr){5-7}
           & PSNR $\uparrow$ & SSIM $\uparrow$ & LPIPS $\downarrow$
           & PSNR $\uparrow$ & SSIM $\uparrow$ & LPIPS $\downarrow$ \\
    \midrule
    Baseline   & 23.28 & 0.812 & 0.217 & 25.47 & 0.879 & 0.147 \\
    +control   & 23.36 & 0.816 & 0.214 & 25.55 & 0.879 & 0.147 \\
    +control \& split  & 23.47 & 0.818 & 0.199 & 25.74 & 0.880 & 0.145 \\
    +control \& split \& structure: (Ours)      & \textbf{23.66} & \textbf{0.822} & \textbf{0.196} & \textbf{26.72} & \textbf{0.884} & \textbf{0.141} \\
    \bottomrule
  \end{tabular}
\end{table}

\begin{figure}[h]
  \centering
  \includegraphics[width=0.7\linewidth]{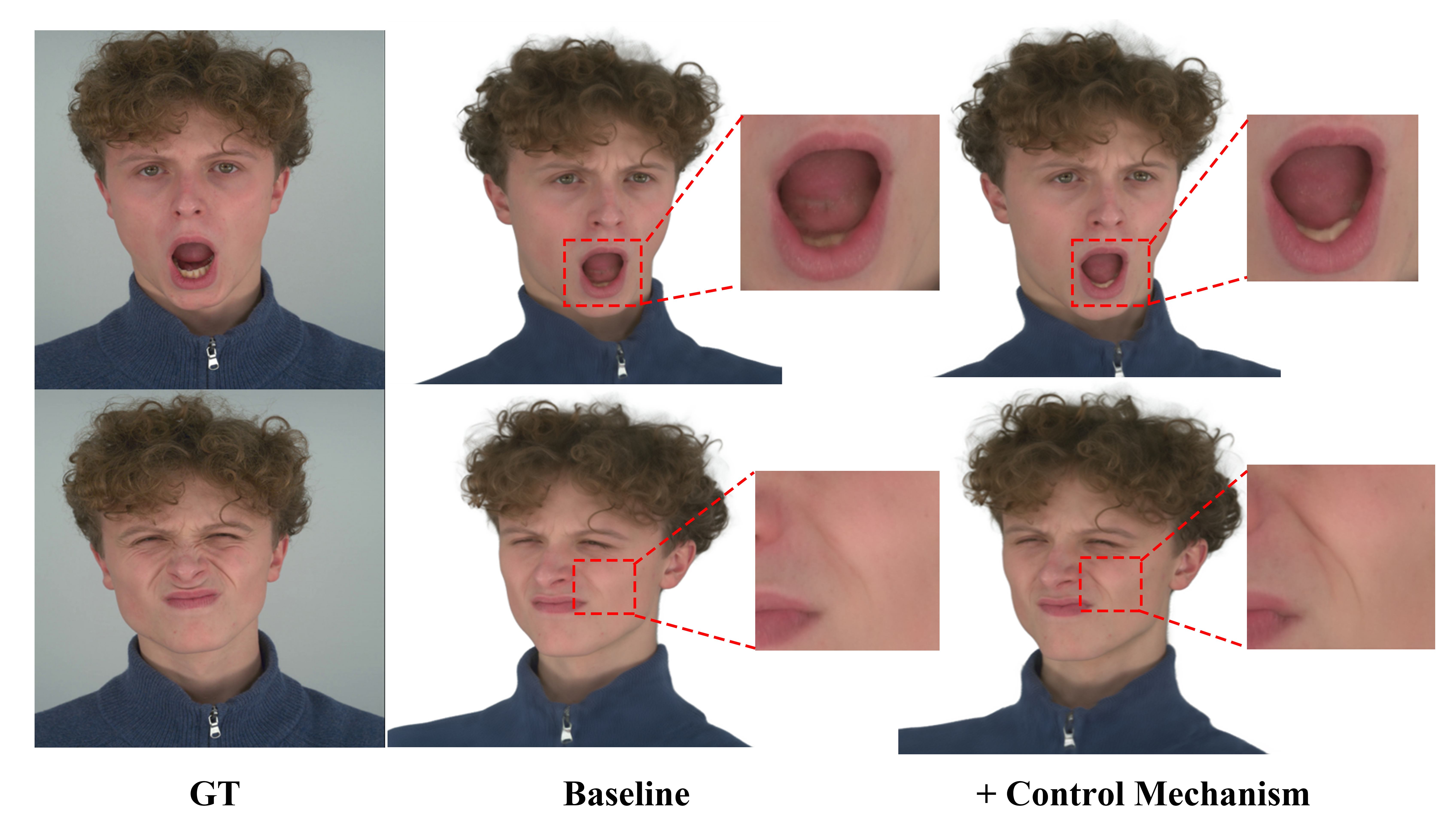}
  \caption{Ablation study on controllable Gaussian mechanism. The figure shows the results of self-reenactment task.}
  \label{fig:self_ablation}
  
\end{figure}

\subsubsection{Controllable Gaussian Mechanism.} To verify the effectiveness of our proposed controllable Gaussian mechanism for modeling expressive details, we compare it with a baseline that does not use control point propagation. In the baseline setting, all Gaussian attributes are directly predicted by the MLPs, without any local structural constraints or guidance. In contrast, our method explicitly selects Gaussians with large predicted displacements as control points and applies a neighborhood propagation strategy to adjust nearby Gaussians. This mechanism improves the local continuity and geometric consistency, especially in regions undergoing strong facial deformation.

\begin{figure}[h]
  \centering
  \includegraphics[width=0.7\linewidth]{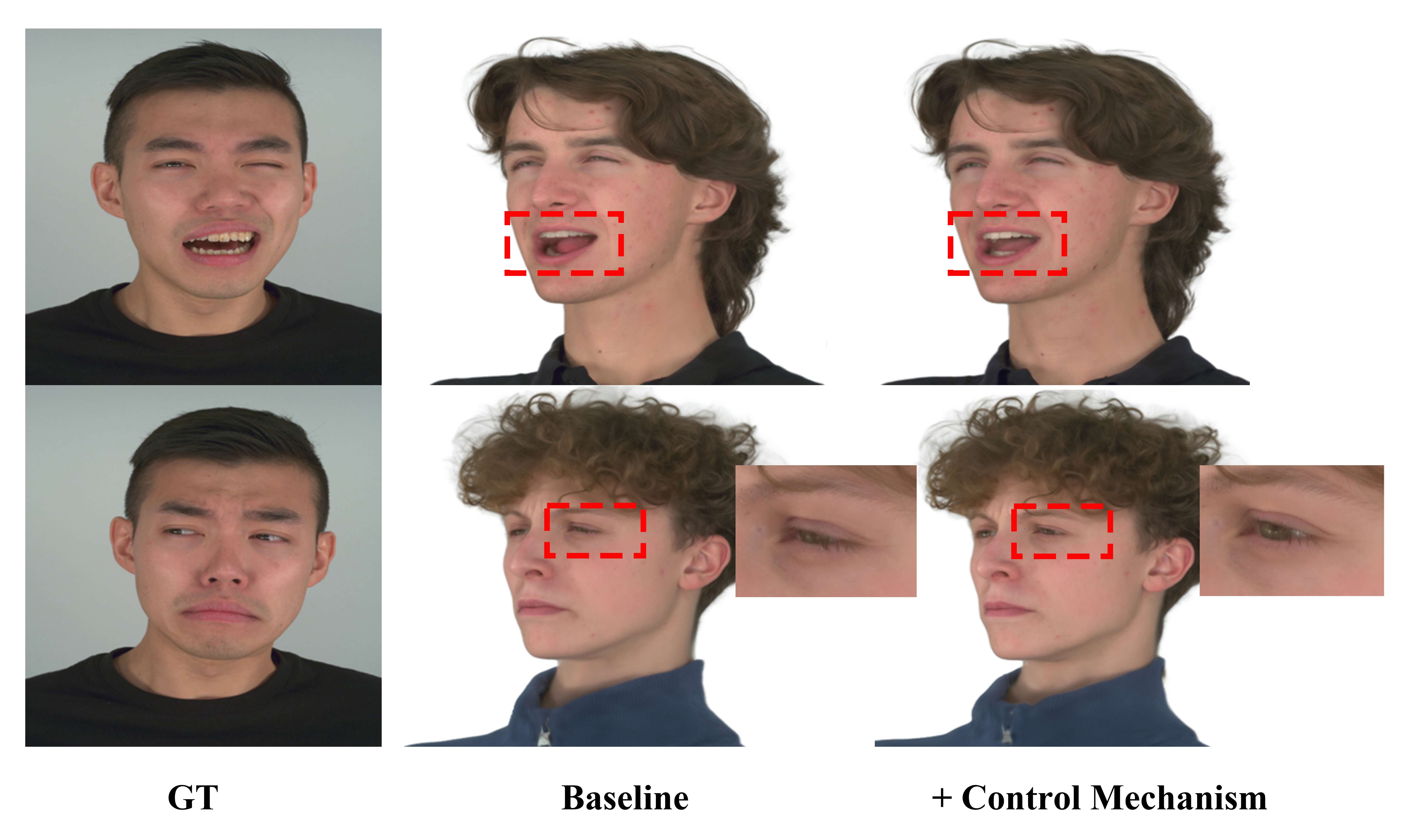}
  \caption{Ablation study of the controllable Gaussian mechanism on the cross-identity reenactment task.}
    \label{fig:cross_ablation}
\end{figure}
We show the results of the visualisation comparison on the self-reenactment and cross-identity Reenactment tasks in Fig~\ref{fig:self_ablation} and Fig~\ref{fig:cross_ablation}, respectively. As shown in these two figures, under expressions such as mouth opening or eyebrow raising, the baseline model tends to produce geometric distortions and blurry artifacts around key regions like the mouth corners and eyes. In comparison, our method generates clearer boundaries and more consistent local details by explicitly guiding deformation through the proposed control point propagation. We also performed a comparison of the reconstruction accuracy on the self-reenactment task for validation, refer to Table~\ref{tab:ablation-results}.

\subsubsection{Splitting Strategy.} Compared to opacity-based splitting strategies, our approach determines Gaussian splitting based on the degree of attribute displacement. Specifically, when a Gaussian undergoes significant shifts in its properties, we trigger a split. This strategy enables our method to more faithfully capture fine-grained local variations, especially those that occur during dynamic expression changes. 

\begin{figure}[h]
  \centering
  \includegraphics[width=0.5\linewidth]{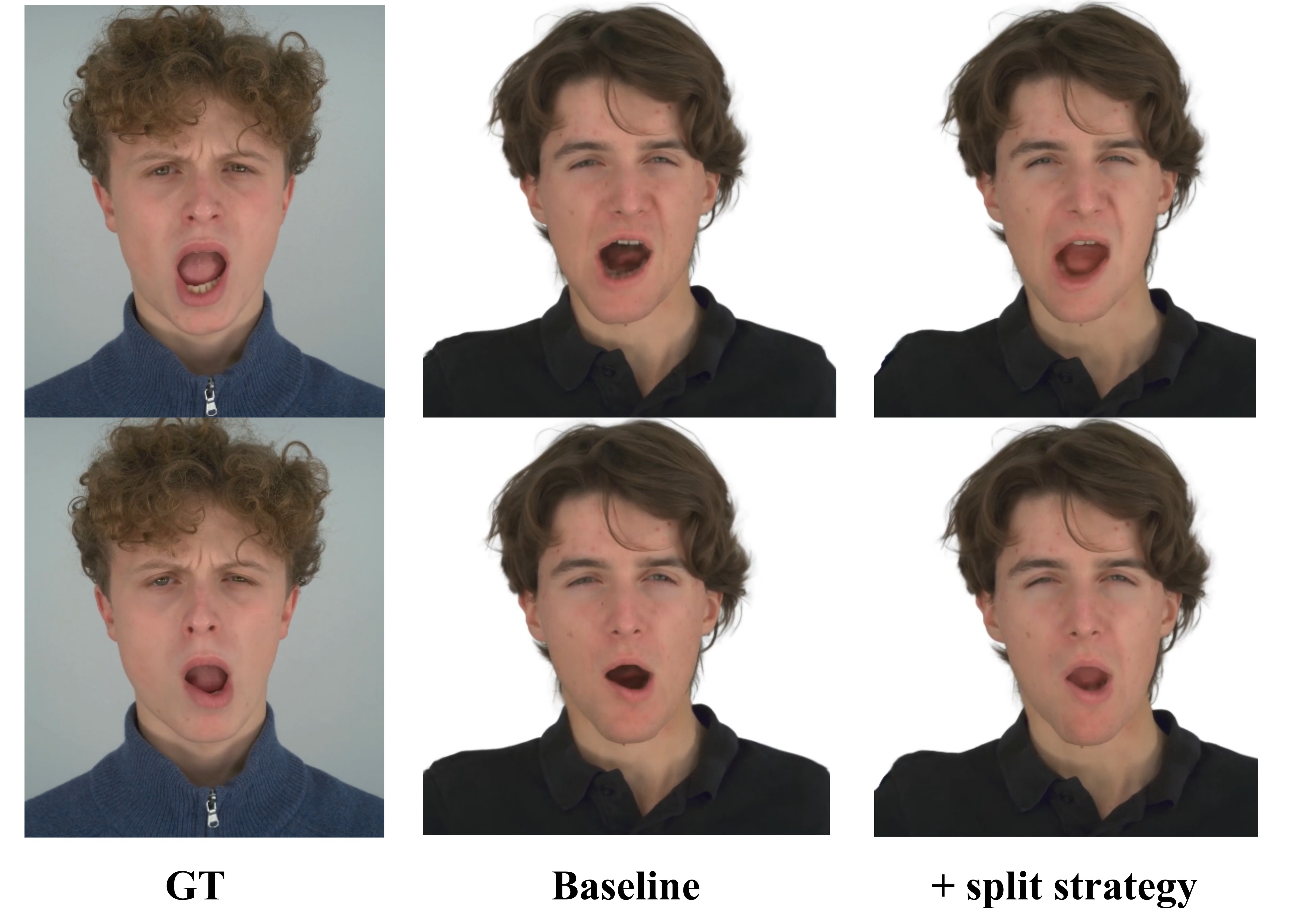}
  \caption{Ablation study of the split strategy on the cross-identity reenactment task.}
\label{fig:split_ablation}
\end{figure}

Fig~\ref{fig:split_ablation} presents a comparison between our splitting strategy and the baseline method, which relies solely on opacity to guide the splitting process. As shown in the figure, by tracking significant changes in Gaussian attributes, our method effectively adapts to local deformations and produces more accurate and natural reconstructions under large expression changes. For example, the appearance of the inner mouth and teeth is rendered more realistically and naturally.

\subsubsection{Additional Ablation on Stage I Objective}

To further analyze the role of the Stage I objective, we provide additional ablations on the loss terms used in $L_{\text{init}}$. Since the objective contains several terms with different functions, we organize the ablation by functionality rather than removing every term independently. Note that the effect of the prior-guided mesh alignment term $L_{\text{mesh}}$ has already been separately validated in the main manuscript through the ablation of the structure-aware modeling strategy in Sec.~4.5.1, where removing the generative-prior-guided geometry modeling degrades both geometric quality and final reconstruction performance. Therefore, the supplementary ablation here focuses on the remaining Stage I losses.

\begin{figure}[H]
    \centering
    \includegraphics[width=0.9\linewidth]{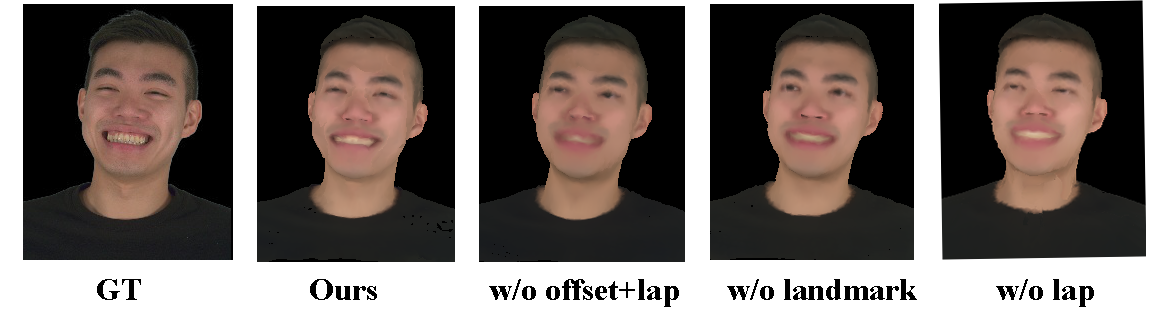}
    \caption{Visual comparison of Stage I loss ablations after the same 10 training epochs. From left to right: ground truth, full model, w/o $\{L_{\text{offset}}, L_{\text{lap}}\}$, w/o $L_{\text{lmk}}$, and w/o $L_{\text{lap}}$.}
    \label{fig:stage1}
\end{figure}
Specifically, we consider: (A) the observation-alignment constraints $\{L_{\text{sil}}, L_{\text{lmk}}\}$; (B) the regularization pair $\{L_{\text{offset}}, L_{\text{lap}}\}$; (C) removing $L_{\text{lmk}}$ only; and (D) removing $L_{\text{lap}}$ only. In practice, we report ablations for Groups B, C, and D, while keeping the remaining terms unchanged.

For Group A, removing $\{L_{\text{sil}}, L_{\text{lmk}}\}$ causes Stage I training to fail at a very early stage, before a valid mesh can be formed. In practice, the optimization collapses and further leads to empty intermediate inputs in the color prediction branch, which prevents training from proceeding normally. This indicates that the observation-alignment constraints are essential not only for improving initialization quality, but also for maintaining the feasibility and stability of Stage I optimization.

\begin{table}[H]
  \centering
  \caption{Quantitative comparison of different Stage I loss ablations on the self-reenactment task (Case 1). The metrics are reported on the final reconstruction after completing Stage II.}
  \label{tab:ablation-stage1}
  \small
  \begin{tabular}{lccc}
    \toprule
    Method & PSNR $\uparrow$ & SSIM $\uparrow$ & LPIPS $\downarrow$ \\
    \midrule
    Ours           & \textbf{24.82} & \textbf{0.903} & \textbf{0.141} \\
    w/o offset+lap & 24.67          & 0.896          & 0.145          \\
    w/o landmark   & 24.71          & 0.902          & 0.144          \\
    w/o lap        & 24.78          & 0.900          & 0.142          \\
    \bottomrule
  \end{tabular}
\end{table}

As shown in Fig.~\ref{fig:stage1}, the Stage I visual results after the same 10 training epochs already reveal clear differences among the ablation settings. Removing $\{L_{\text{offset}}, L_{\text{lap}}\}$ leads to the most obvious degradation in local facial geometry, producing blurrier and less accurate reconstruction in expressive regions. Removing $L_{\text{lmk}}$ weakens structural alignment, while removing $L_{\text{lap}}$ reduces local smoothness and geometric stability. To further verify the impact on the final reconstruction quality, we also report quantitative results after completing Stage II in Table~\ref{tab:ablation-stage1}. The full model consistently achieves the best performance, while the ablated variants all show degraded PSNR, SSIM, or LPIPS, which is consistent with the visual observations from Stage I.

Overall, these results show that the Stage I loss terms play complementary roles. The observation-alignment constraints $\{L_{\text{sil}}, L_{\text{lmk}}\}$ are necessary for stable optimization, while $L_{\text{offset}}$ and $L_{\text{lap}}$ further improve geometric accuracy and regularity. The role of $L_{\text{mesh}}$ is separately validated in Sec.~4.5.1 through the ablation of the structure-aware modeling strategy.

\subsubsection{Threshold Selection and Analysis.} 
\label{sec:threshold} Although facial geometries differ across individuals, we observe that the degree and pattern of expression-induced deformation, relative to the neutral face, remain largely consistent, particularly in active facial regions such as the mouth, eyes, and brows. This indicates that the displacement magnitude and variation trend of expressive areas are generally similar across identities, supporting the feasibility of a unified control strategy.

We further analyzed the typical amplitude of motion in expressive regions and their spatial coverage across identities, and found them to consistently fall within moderate ranges. This observation informed our choice of a control threshold of 0.3, which effectively captures significant expression-driven displacements while avoiding over-propagation. Similarly, a splitting threshold of 0.2 was found to provide early refinement in moderately active regions without introducing unnecessary growth. These values align with early-stage experiments and were fixed throughout all evaluations.

To further validate the choice of the control threshold, we conduct quantitative comparisons across different values on representative cases.
As shown in Table~\ref{tab:control-threshold}, a threshold of 0.3 achieves the highest PSNR and SSIM, while lower values such as 0.15 tend to over-propagate motion, and higher values such as 0.4 fail to capture subtle deformations.

\begin{table}[ht]
  \centering
  \caption{Impact of different control thresholds on two representative cases.}
  \label{tab:control-threshold}
  \small
  \begin{tabular}{lcccc}
    \toprule
    Threshold & \multicolumn{2}{c}{Case 1} & \multicolumn{2}{c}{Case 2} \\
    \cmidrule(lr){2-3} \cmidrule(lr){4-5}
              & PSNR $\uparrow$ & SSIM $\uparrow$ & PSNR $\uparrow$ & SSIM $\uparrow$ \\
    \midrule
    0.15        & 23.86 & 0.795 & 24.72 & 0.858 \\
    \textbf{0.30} & \textbf{24.24} & \textbf{0.814} & \textbf{24.82} & \textbf{0.903} \\
    0.40        & 24.01 & 0.796 & 24.48 & 0.901 \\
    \bottomrule
  \end{tabular}
\end{table}

\begin{figure}[h]
  \centering
  \includegraphics[width=0.8\linewidth]{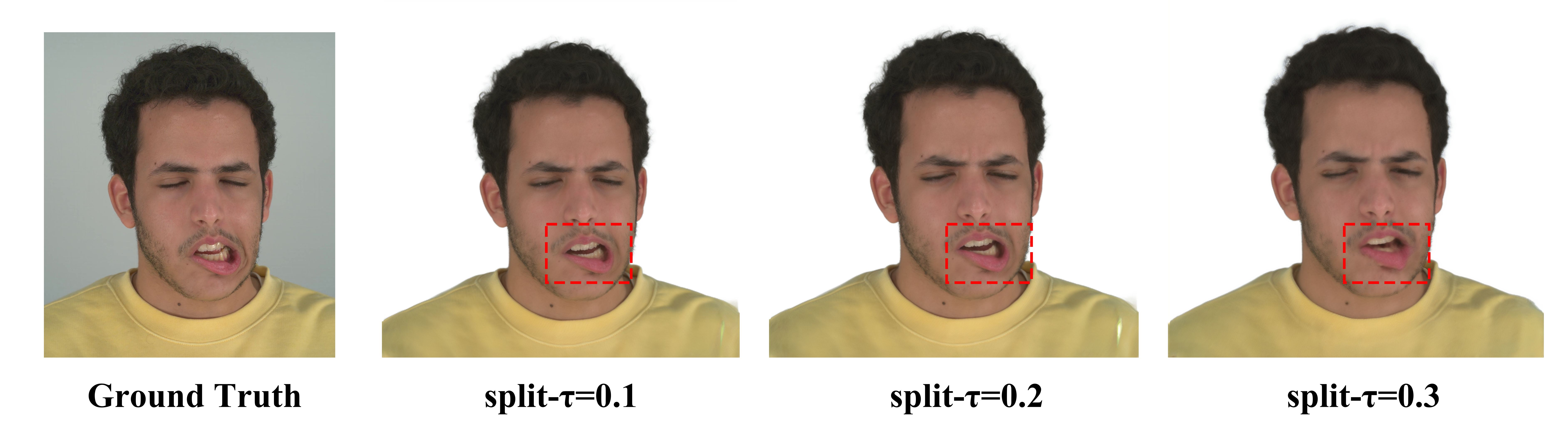}
  \caption{Ablation study on splitting threshold. The figure shows the results of self-reenactment task.}
  \label{fig:split-ablation}
\end{figure}

Unlike the control threshold, the splitting threshold primarily governs the number of regions selected for refinement at each iteration, thereby controlling the number of newly introduced Gaussians with the aim of capturing fine-grained local details. A lower threshold (e.g., 0.1) triggers more frequent splitting operations, resulting in nearly twice as many new Gaussians per iteration compared to a threshold of 0.2. However, as shown in Fig~\ref{fig:split-ablation}, it brings almost no perceptual improvement over the 0.2 threshold. In contrast, a higher threshold (e.g., 0.3) delays region updates and fails to timely capture changes in visual details, especially around subtle facial regions such as lips and teeth. 

\subsubsection{Quantitative Analysis.} To comprehensively evaluate the contribution of each proposed module, we conduct an ablation study starting from a baseline that excludes expression control and structure-aware modeling strategy. Table~\ref{tab:ablation-results} compares different combinations by progressively adding the expression control mechanism, the Gaussian splitting strategy, and the structure-aware modeling strategy. Overall, we observe consistent improvements across all metrics—PSNR, SSIM, and LPIPS—as modules are gradually introduced. Specifically, incorporating the expression control mechanism leads to a notable gain in PSNR, indicating better reconstruction of overall image structure. While SSIM and LPIPS show smaller or slightly fluctuating changes in some cases, the model remains stable in performance. Nevertheless, these metrics may overlook certain perceptual aspects. As shown in Fig~\ref{fig:self_ablation} and Fig~\ref{fig:cross_ablation}, our method yields consistent qualitative improvements—especially in mouth and eye movement as well as teeth visibility—revealing subtle yet meaningful changes that go beyond what standard perceptual metrics can capture. Finally, with the addition of the structure-aware geometry modeling module, our method achieves the best results across all three metrics, demonstrating the strong synergy among the proposed components.

\subsubsection{Efficiency Analysis}

Regarding efficiency, we note that the Signed Distance Function (SDF) is used as a geometry prior to stabilize the early phase of Gaussian optimization, and this process is executed only once per subject. The full training pipeline, including SDF-based initialization, Gaussian attribute learning, and expression control optimization, takes approximately 2.5 days per subject on a single NVIDIA RTX 3090 GPU, as summarized in Table~\ref{tab:training_rendering}. Although some recent 3DGS-based methods achieve shorter training time and higher rendering FPS, our method is designed to provide a different trade-off point by prioritizing reconstruction fidelity and controllability under more challenging settings. 

\begin{table}[ht]
  \centering
  \caption{Comparison of training time and rendering speed across methods.}
  \label{tab:training_rendering}
  \small
  \begin{tabular}{lcc}
    \toprule
    Method & Training Time $\downarrow$ & Rendering FPS $\uparrow$ \\
    \midrule
    NeRFace  & 4 days   & \string~
0.06 \\
    HAvatar   & 7 days   & \string~
7 \\
    GHA            & 3 days   & \string~
32 \\
    GaussianAvatars  & 1 day &  \string~ 183\\
    TensorialGA      &1 day & \string~ 150\\
    Ours                             & \textbf{2.5 days} & \textbf{\string~
32} \\
    \bottomrule
  \end{tabular}
\end{table}

At inference time, our method adopts rasterization-based rendering of 3D Gaussians and achieves approximately 32 FPS on an RTX 3090, which still supports smooth rendering for interactive or near real-time applications.

To further examine the trade-off between rendering efficiency and reconstruction quality, we provide an additional comparison with GaussianAvatars, a representative recent 3DGS-based head avatar method with substantially higher rendering speed. As summarized in Table~\ref{tab:ga_tradeoff}, GaussianAvatars achieves much higher FPS, while our method obtains consistently stronger reconstruction quality under the unified and more challenging evaluation setting used in this work.

It is worth noting that the efficiency advantage of GaussianAvatars is associated with its lightweight FLAME-rigged Gaussian design, in which each Gaussian is directly attached to a FLAME triangle and rendered through an efficient rasterization pipeline. At the same time, this design also relies on a relatively complex preprocessing procedure. In our implementation, the average preprocessing time of GaussianAvatars is substantially longer than that of our method.

\begin{table}[H]
\centering
\caption{Efficiency--performance trade-off comparison with GaussianAvatars. Results are averaged across subjects under our more challenging evaluation setting.}
\label{tab:ga_tradeoff}
\small
\setlength{\tabcolsep}{4pt}
\begin{tabular}{lccccc}
\toprule
Method & Prep. (h)$\downarrow$ & FPS$\uparrow$ & PSNR$\uparrow$ & SSIM$\uparrow$ & LPIPS$\downarrow$ \\
\midrule
GaussianAvatars & 32.7  & \textbf{183} & 22.50 & 0.826 & 0.207 \\
Ours            & \textbf{12.0} & 32 & \textbf{25.07} & \textbf{0.857} & \textbf{0.168} \\
\bottomrule
\end{tabular}
\end{table}

In addition, compared with the original protocol of GaussianAvatars, our evaluation setting is more challenging because all prescribed sequences are used for training and the FREE sequence is reserved for testing, resulting in a larger train--test gap and higher motion diversity. Under this setting, GaussianAvatars still provides substantially higher rendering efficiency, whereas our method achieves stronger reconstruction quality. These results suggest that the two methods correspond to different operating points on the efficiency-performance spectrum, and that our method provides a favorable trade-off when reconstruction fidelity under more challenging conditions is prioritized.

\section{Spatial Distribution of Control Gaussians}

\begin{figure}[H]
    \centering
    \includegraphics[width=0.6\linewidth]{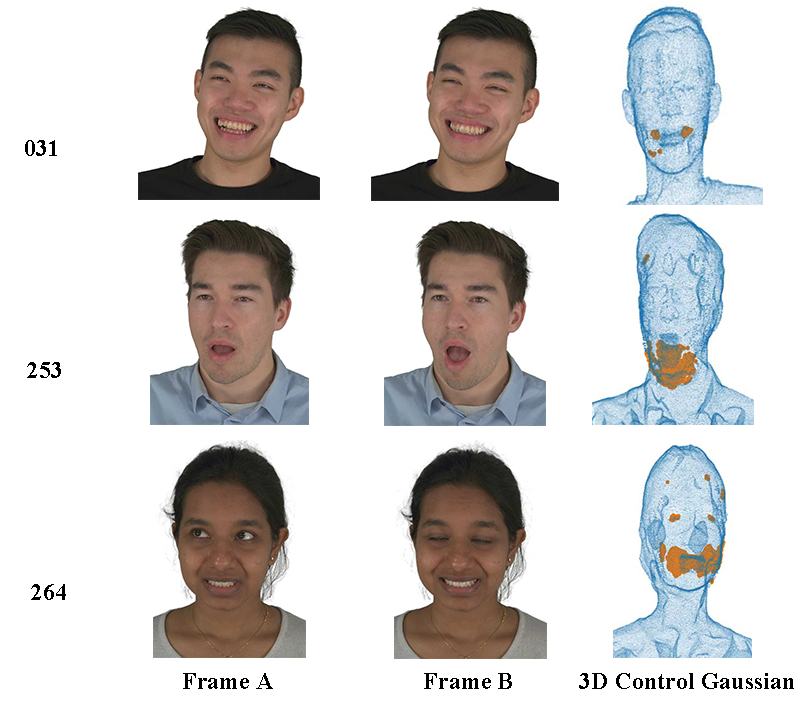}
    \caption{Visualization of the selected Control Gaussians for different subjects. Each row shows two representative frames with clear expression variation and the corresponding 3D distribution of the selected Control Gaussians. The highlighted points consistently concentrate in semantically expressive and highly deformable facial regions, while still showing subject-specific variation.}
    \label{fig:control}
\end{figure}

To better understand where the selected Control Gaussians lie in the reconstructed 3D volume, we visualize their spatial distribution for multiple subjects in Fig.~\ref{fig:control}. Each example includes two representative frames with clear expression variation, together with the corresponding 3D distribution of the selected Control Gaussians.

As shown in Fig.~\ref{fig:control}, the selected Control Gaussians are not randomly distributed, but consistently concentrate in semantically expressive facial regions, such as the periocular area, the mouth, and the lower face. At the same time, the dominant activated regions are not exactly identical across subjects, indicating that the proposed control selection mechanism is semantically meaningful while remaining adaptive to subject-specific deformation patterns.

It is important to note that the correspondence between visible 2D appearance changes and the selected 3D Control Gaussians should be understood at the semantic-region level, rather than as a strict point-to-point spatial match. In general, the selected Control Gaussians tend to appear in highly deformable facial regions. However, in our method, they are selected according to the learned displacement response, rather than by explicit mesh reconstruction residuals or predefined anatomical priors. Therefore, while they often overlap with regions undergoing large deformations, they should not be interpreted as a direct indicator of where the neutral mesh is poorly reconstructed.

\section{Limitations and Future Work}
Although our method improves local expression control and geometric stability for dynamic head avatar reconstruction, some challenges remain. In particular, highly articulated or partially occluded local regions, especially the mouth interior, are still difficult to model robustly due to their complex topology, frequent visibility changes, and relatively weak effective supervision. In addition, because our framework is primarily designed for head avatar reconstruction, neighboring non-facial regions and challenging pose or illumination variations are not modeled as explicitly as facial dynamics, which may lead to reduced stability in some cases. These limitations are also common in dynamic head avatar reconstruction. Future improvements may therefore come from stronger structural priors for mouth interior reconstruction, more explicit temporal stabilization for neighboring non-facial regions, explicit eye-pose modeling for finer controllability, and illumination-aware avatar representations for improved robustness under large pose and lighting variations.

\section{Conclusion}
In this paper, we propose an expression-aware and deformation-aware 3D avatar reconstruction framework leveraging dynamic 3D Gaussian representations. By introducing a controllable Gaussian mechanism and a deformation-aware splitting strategy, our method improves geometric expressiveness and local consistency in regions with complex expression changes. Additionally, we introduce a structure-aware geometry modeling module guided by a pretrained large-scale avatar prior, which significantly improves geometric stability and structural consistency during training. Experimental results show that our approach outperforms existing methods across multiple metrics, enabling high-quality synthesis and more realistic expression-driven avatar reconstruction.

\bibliographystyle{ACM-Reference-Format}
\bibliography{sample-base}


\end{document}